\documentclass[letterpaper]{article} 
\usepackage{aaai24}  
\usepackage{times}  
\usepackage{helvet}  
\usepackage{courier}  
\usepackage[hyphens]{url}  
\usepackage{graphicx} 
\usepackage{natbib}  
\usepackage{caption} 
\usepackage{algorithm}
\usepackage{algorithmic}

\usepackage{newfloat}
\usepackage{listings}
\DeclareCaptionStyle{ruled}{labelfont=normalfont,labelsep=colon,strut=off} 
\floatstyle{ruled}
\newfloat{listing}{tb}{lst}{}
\floatname{listing}{Listing}
\usepackage{threeparttable} 
\usepackage{booktabs,makecell,multirow} 
\usepackage{bbding}
\usepackage{color, soul, colortbl}
\usepackage[dvipsnames]{xcolor}
\usepackage{verbatim}
\usepackage{amsfonts, amssymb, amsmath, bm}
\usepackage{subfigure}

\title{MICA: Towards Explainable Skin Lesion Diagnosis via Multi-Level Image-Concept Alignment}
\author{
    Yequan Bie\textsuperscript{\rm 1}, Luyang Luo\textsuperscript{\rm 1}, Hao Chen\textsuperscript{\rm 1,2,3}\thanks{Corresponding author.}
}
\affiliations{
    \textsuperscript{\rm 1}Department of Computer Science and Engineering, Hong Kong University of Science and Technology\\
    \textsuperscript{\rm 2}Department of Chemical and Biological Engineering, Hong Kong University of Science and Technology\\
    \textsuperscript{\rm 3}HKUST Shenzhen-Hong Kong Collaborative Innovation Research Institute\\
    ybie@connect.ust.hk, cseluyang@ust.hk, jhc@cse.ust.hk\\

}

\usepackage{bibentry}

\begin{document}

\maketitle

\begin{abstract}

Black-box deep learning approaches have showcased significant potential in the realm of medical image analysis. However, the stringent trustworthiness requirements intrinsic to the medical field have catalyzed research into the utilization of Explainable Artificial Intelligence (XAI), with a particular focus on concept-based methods. Existing concept-based methods predominantly apply concept annotations from a single perspective (e.g., global level), neglecting the nuanced semantic relationships between sub-regions and concepts embedded within medical images. This leads to underutilization of the valuable medical information and may cause models to fall short in harmoniously balancing interpretability and performance when employing inherently interpretable architectures such as Concept Bottlenecks. To mitigate these shortcomings, we propose a multi-modal explainable disease diagnosis framework that meticulously aligns medical images and clinical-related concepts semantically at multiple strata, encompassing the image level, token level, and concept level. Moreover, our method allows for model intervention and offers both textual and visual explanations in terms of human-interpretable concepts. Experimental results on three skin image datasets demonstrate that our method, while preserving model interpretability, attains high performance and label efficiency for concept detection and disease diagnosis. The code is available at \url{https://github.com/Tommy-Bie/MICA}.

\end{abstract}

\section{1 \quad Introduction}

Black-box deep learning methods have surfaced as powerful instruments in medical image analysis, offering significant potential to revolutionize healthcare diagnostics and treatments \cite{kermany2018identifying,esteva2017dermatologist}. These methods excel at handling the extensive and intricate data inherent to the medical field, 
rendering them suitable for many tasks \cite{litjens2017survey}. Despite the encouraging performance, their end-to-end prediction nature leads to a lack of transparency, raising critical issues of trust and interpretability in high-stakes domains like healthcare \cite{rudin2019stop}.


\begin{figure}[t]
\centering 
\includegraphics[width=0.85\linewidth, height=4cm]{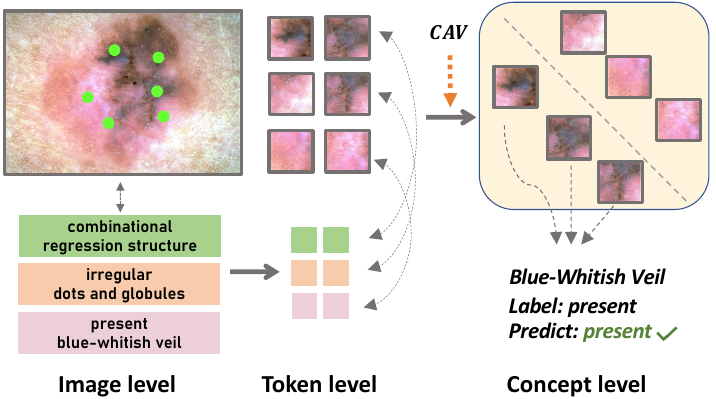} 
\caption{Our method learns image and concept semantic correspondences at the image, token, and concept levels.} 
\label{multi_level}
\end{figure}

The healthcare field, with its rigorous demands for trustworthiness, requires models that not only perform well but are also understandable and trustable by practitioners, which necessitates research into Explainable Artificial Intelligence (XAI). Within XAI, several approaches have been proposed to explain the neural networks using saliency map \cite{cam,gradcam} highlighting the contribution of each pixel or region in the model's prediction, while others leverage inspection of the learned features \cite{abbasi2017structural}, feature interactions \cite{tsang2017detecting}, and influence functions \cite{koh2017understanding} to explain the models. However, the reliability of these post-hoc analyses, which offer explanations for a trained AI model, has been under considerable scrutiny recently. Some studies \cite{laugel2019dangers,rudin2019stop} have shown that post-hoc explanation techniques often yield inconsistent results across different runs and are sensitive to slight changes in the input, making the post-hoc methods misleading as they could provide explanations that do not accurately reflect the model's decision-making process. 

Thus, ante-hoc explainable methods have garnered researchers' interest, with a particular emphasis placed on concept-based methods. These methods aim to integrate interpretability into machine learning models by linking their predictions to human-understandable concepts \cite{cbm,yuksekgonul2023posthoc,fang2020concept, yan2023towards}. For example, the Concept Bottleneck Model (CBM) \cite{cbm} first predicts an intermediate set of pre-defined concepts and then uses these concepts to predict the final output. \citeauthor{yan2023towards} \shortcite{yan2023towards} introduced a human-in-the-loop framework to eliminate confounding factors and improve model performance. These inherently interpretable methods offer concept-based explanations, which are generally more understandable than post-hoc approaches.

However, concept-based methods are not devoid of limitations. A major challenge these methods face is that the model performance (e.g., classification accuracy) is sacrificed when designing an explainable architecture like Concept Bottlenecks, compared to the black-box methods. We argue this is caused by inefficient usage of valuable medical information, i.e., clinical-related concepts.
Most existing methods apply concept annotations at a single level, e.g., \citeauthor{Sarkar_2022_CVPR} \shortcite{Sarkar_2022_CVPR} only utilize concept labels to supervise the concept prediction results of the whole image, neglecting the intricate semantic connections between images' sub-regions and concepts. 
This narrow focus can limit the models' performance and interpretability, leading to unreliable concept detection and inaccurate diagnosis. 

To address the mentioned challenges, we introduce a multi-modal explainable disease diagnosis framework to meticulously align medical images and clinical-related concepts semantically at multiple levels, encompassing the global image level, the regional token level, and the concept subspace level as shown in Figure \ref{multi_level}. Specifically, image-level alignment encourages the model to learn the correspondences between images and concepts from a global perspective. Token-level alignment focuses on the similarity between sub-regions within images and concept tokens using an attention-based mechanism. Concept-level alignment leverages concept activation vectors (CAVs) \cite{TCAV} to project the concept-based attention-weighted image representations to a concept subspace and subsequently aligns the aggregated image representations with clinical concepts. It is noteworthy that since the utilized concepts are human-interpretable, we leverage the knowledge of a medical large language model (LLM) by employing it as a concept encoder to enable the model to comprehend the latent conceptual semantics. During disease diagnosis, our model detects the concepts before making the decision. In this manner, our method makes full use of concept-based medical semantics through multi-level image-concept joint learning and achieves better performance and interpretability.

We summarize our main contributions as follows: (1) We propose MICA, a novel explainable disease diagnosis framework that semantically aligns medical images and clinical concepts at three different levels, i.e., global image level, regional token level, and concept subspace level. (2) To the best of our knowledge, we are the first to encode dermoscopic concepts using medical LLM within XAI. (3) As an ante-hoc explainable framework, our method is capable of performing disease diagnosis and concept detection concurrently while offering both visual and textual explanations. (4) Experimental results on three skin datasets show that our method achieves superior performance and label efficiency, benefiting from the high-quality semantic correlations between images and concepts learned within our framework. 

\section{2 \quad Related Works}
\subsection{2.1 \quad XAI \& Concept-based Methods}
With an increasing number of high-stake scenarios (e.g., healthcare, finance, law enforcement) requiring trustworthiness, XAI has been gaining attraction. One general approach in XAI is post-hoc analysis, which aims to interpret a trained model by fitting explanations to the model outputs, such as LIME \cite{lime}, SHAP \cite{shap}, and SENN \cite{senn}. Particularly for CNN, many researchers focus on saliency visualization \cite{cam,gradcam,axiomatic} and activation maximization \cite{van2016pixel,yosinski2015understanding,nguyen2016synthesizing}. However, post-hoc methods, which typically provide explanations based on pixels, regions or features of input images, do not genuinely enable medical experts or patients to understand which specific symptoms contribute to the decision process. This premise has sparked researchers' interest in the exploration of concept-based methods that integrate high-level human-interpretable concepts into decision process. Several researchers work on automatically discovering the concepts \cite{auto1,auto2}, which can reduce the need for concept annotations but may not be suitable for healthcare, since the semantic meanings of discovered concepts can be unclear and unreliable. Concept activation vectors (CAVs) like approaches \cite{TCAV, lucieri2020interpretability, Patricio_2023_CVPR, yan2023towards} train linear classifiers, e.g., SVM \cite{svm}, to the model's features to verify whether the representations can separate the human-defined concept examples. The inherently interpretable Concept Bottleneck Model \cite{cbm, Patricio_2023_CVPR, rigotti2021attention} first predicts concepts, then uses the detected concepts to predict task labels. We argue CBM is an essential research direction in trustworthy medical image analysis, since it mimics the process wherein medical experts first assess symptoms before diagnosing diseases during clinical treatments.


\begin{figure*}[t]
\centering
\includegraphics[width=2.0\columnwidth, height=9cm]{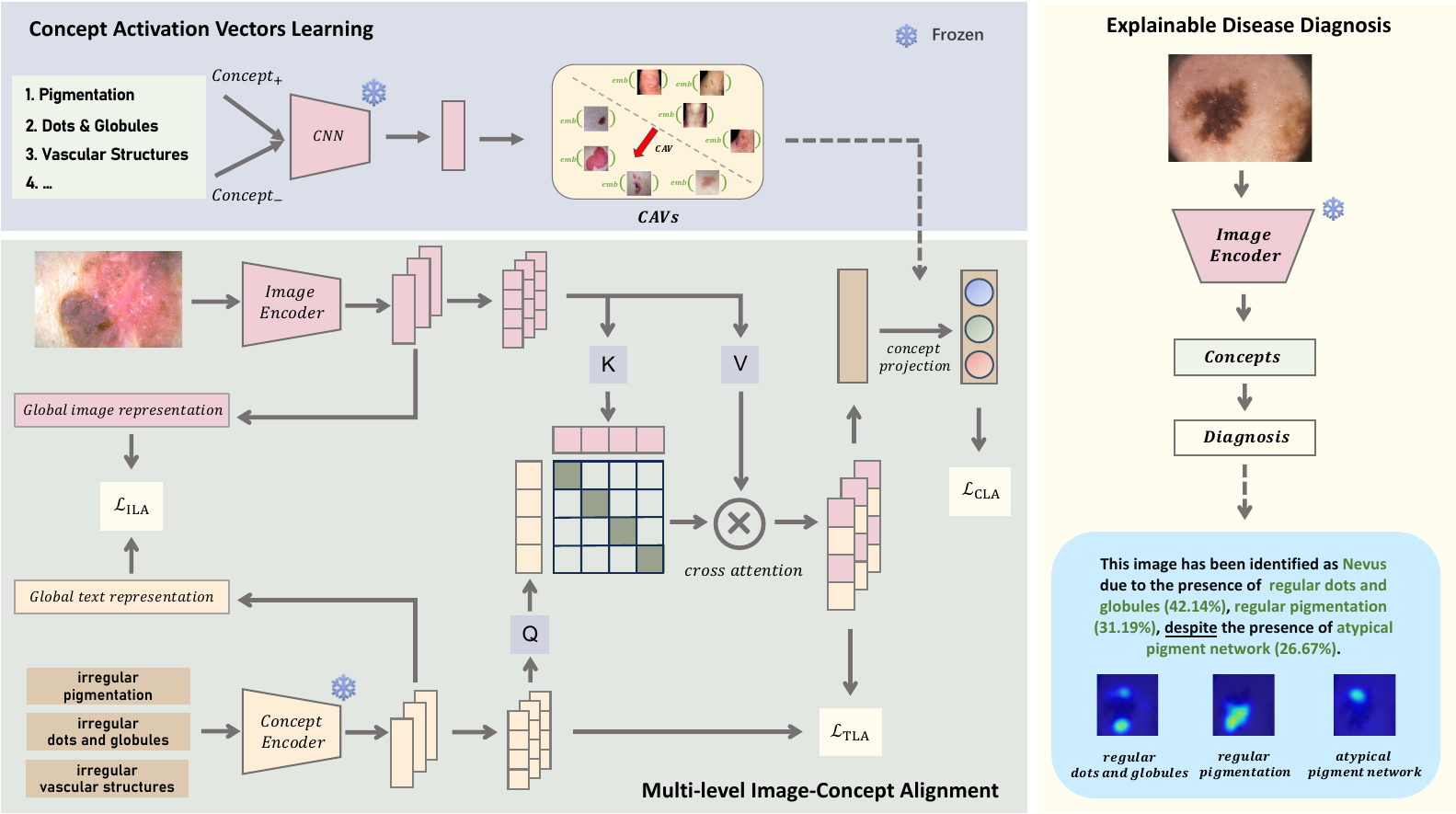} 
\caption{The overall pipeline of our proposed framework.}
\label{overview}
\end{figure*}

\subsection{2.2 \quad Trustworthy Skin Disease Diagnosis}
The diagnosis of skin diseases, especially skin cancer, has been a significant research area within the intersection of deep learning and healthcare. Many of explanation approaches for skin lesion diagnosis are based on saliency map \cite{young2019deep,xiang2019towards} and attention mechanisms \cite{barata2021explainable,canet}. However, considering the stringent demands for model decision interpretability in healthcare \cite{lipton2017doctor}, some researchers have taken efforts in designing concept-based models upon the ABCD-rule \cite{nachbar1994abcd} and the 7-point checklist \cite{7pt_checklist}, which are authoritative criteria established by dermatologists. For instance, Lucieri et al. \shortcite{lucieri2020interpretability} predict dermoscopic concepts from a pre-trained network to explain its predictions using TCAV. Coppola et al. \shortcite{mtl} propose to predict dermoscopic features with information sharing between different subnetworks to increase interpretability through multi-task learning. Yan et al. \shortcite{yan2023towards} discover and eliminate confounding concepts within the datasets using spectral relevance analysis \cite{lapuschkin2019unmasking}. CBE \cite{Patricio_2023_CVPR} uses an extra segmentation module to preprocess images and encourages the feature maps obtained by $1\times1$ convolutional kernels to learn the representations of each dermoscopic concept, then diagnoses diseases in CBM architecture. However, most existing methods primarily utilize concept annotations from a single perspective and focus on the analysis of dermoscopic images. In contrast, our method semantically aligns skin images and concepts at multiple levels without using extra models to obtain disease masks, which can be applied to both dermoscopic images and raw, clinical images, e.g., SkinCon dataset \cite{daneshjou2022skincon}. 

\section{3 \quad Method}

\subsection{3.1 \quad Overall Framework}
Figure \ref{overview} presents the overall architecture of our multi-level image-concept alignment framework for explainable disease diagnosis. Our method mainly consists of two stages: multimodal representation learning through image-concept alignment and explainable disease diagnosis. Specifically, in the first stage, we utilize a CNN-based image encoder and a large language model (LLM)-based concept encoder to extract semantic visual and textual features from the input medical images and corresponding clinical-related concepts. Then we align the images and concepts at three levels, i.e., image level, token level and concept level, directing the feature extractor to more effectively leverage the correspondences between images and concepts. 
To elaborate, we employ an image-level alignment module to maximize the similarity of the global representation between accurate image-concept pairs versus random pairs. Then, an attention-based token-level image-concept alignment module is proposed to cultivate fine-grained alignments between image sub-regions and concept word tokens. Moreover, to further refine the image and concept matches established by the first two modules, we introduce a concept-level alignment module based on concept activation vectors \cite{TCAV}, which maps the aggregated attention-weighted image representation onto the concept subspace and subsequently enhances the match with the concept ground truth. 
In the second stage, we add a single layer on top of the image encoder trained in the first stage to predict the concepts, and use the detected concepts to predict the diagnosis through a linear probe. Finally, concept-based explanations, including concept contributions and localization, are generated.


\subsection{3.2 \quad Multi-Level Image-Concept Alignment}
\subsubsection{Image-level Image-concept Alignment.} 
To encourage the model to learn global correspondences between images and concepts, we employ an image-level alignment module. Specifically, given training set $D = \{(I_1, C_1), (I_2, C_2), ..., (I_N, C_N)\}$, where $(I_i, C_i)$ denotes the $i$-th image-concept pairs, and $N$ denotes the number of training samples, we adopt a CNN-based image encoder $E_I: I \rightarrow V$, 
which takes an image $I_i$ as input and outputs the global average pooling result of the last layer's feature map, i.e., the visual representation $V_i$. To extract the semantic representations of concepts, we utilize a medical LLM-based concept encoder $E_C: C \rightarrow \{t, T\}$, 
which outputs the encoded concept text tokens $t_i$ and the aggregated representation $T_i$ given the concept $C_i$. The same dimensional image representation $F_I$ and concept representation $A_I$ are obtained using two projection layers that transform $V$ and $T$ into latent space embeddings, respectively.
Then, the similarity $s_{ij} = (F_{Ii})^T A_{Ij}$ between image representation $F_{Ii}$ and concept representation $A_{Ij}$ is calculated by cosine similarity.
To maximize the alignment between the correct pairs of images and concept versus random pairs, we follow previous work \cite{zhang2022contrastive} to maximize the posterior probability of the image representation given its corresponding concept representation using contrastive loss function:
\begin{equation}
    L_I^{(v|t)} = \sum_{i=1}^N -{\rm{log}}\frac{{{\rm{exp}}(s_{ii}/\tau_1)}}{\sum_{j=1}^N {\rm{exp}}(s_{ij}/\tau_1)},
\end{equation}
where $\tau_1$ is a scaling temperature parameter. Considering the mutual relationship between the image and concept pairs, we also maximize the posterior probability of the concept given its corresponding image by minimizing the symmetrical contrastive loss function:
\begin{equation}
    L_I^{(t|v)} = \sum_{i=1}^N -{\rm{log}}\frac{{{\rm{exp}}(s_{ii}/\tau_1)}}{\sum_{j=1}^N {\rm{exp}}(s_{ji}/\tau_1)}
\end{equation}
\indent The overall objective function of the image-level image-concept alignment module $\mathcal{L}_{\rm{ILA}}$ is the average of the two symmetrical losses.

\subsubsection{Token-level Image-concept Alignment.} In medical imaging, the primary focus is often on specific, small regions within an image. Diagnosis typically hinges on the symptoms observable within these particular image regions, which is different from the natural image domain. Therefore, considering that different clinical concepts may correspond to distinct sub-regions of a medical image, we propose the attention-based token-level image-concept alignment module. This module aims to address the limitation of image-level alignment and additionally explore the correlation between the image features of sub-regions and the respective concept tokens. Given the image encoder $E_I$, we extract the region-level visual feature maps from the intermediate convolution layer and vectorize to get the features of each image sub-region. A projection layer is also applied to the features to get the low-dimensional embeddings $F_T$. Similarly, the concept token features $t$ extracted by concept encoder $E_C$ are projected into token-level text embeddings $A_T$. We then calculate the concept-based attention-weighted image representation $g_i$ using the cross attention between the region-level visual embeddings $F_{Ti}$ and the token-level concept embedding $A_{Ti}$. Specifically, we regard $A_{T}$ as queries, $F_{T}$ as keys and values, then the attention weight $\alpha_{ij}$ is obtained by calculating the following formulation:
\begin{equation}
    \alpha_{ij} = \frac{{\rm{exp}}\left((F_{Ti})^T A_{Tj}/\tau_2\right)}{\sum_{k=1}^M {\rm{exp}}\left((F_{Ti})^T A_{Tk}/\tau_2\right)},
\end{equation}
where $\tau_2$ is a scaling temperature parameter. Then the concept-based attention weighted image representation $g_i = \sum_{j=0}^M\alpha_{ij}F_{Tj}$ is the weighted sum of sub-region features.

In order to align image and concept representations at the token level, we aggregate the similarity between all $W$ concept features and their corresponding attention-weighted image representations using the token-wise matching function:
\begin{equation}
    G(g_i, A_{Ti}) = {\rm{log}}(\sum_{i=1}^W{\rm{exp}}(\langle g_i, A_{Ti}\rangle/\tau_3))^{\tau_3}
\end{equation}
where $\tau_3$ is a scaling temperature parameter. Then, similar to the image-level alignment module, we define the token-level image-concept alignment contrastive losses as:
\begin{align}
    L_T^{(v|t)} &= \sum_{i=1}^N -{\rm{log}}\frac{{{\rm{exp}}(G(g_i, A_{Ti})/\tau_2)}}{\sum_{j=1}^N {\rm{exp}}(G(g_i, A_{Tj})/\tau_2)},\notag \\
    L_T^{(t|v)} &= \sum_{i=1}^N -{\rm{log}}\frac{{{\rm{exp}}(G(g_i, A_{Ti})/\tau_2)}}{\sum_{j=1}^N {\rm{exp}}(G(g_j, A_{Ti})/\tau_2)}
\end{align}
\indent The overall objective of the token-level alignment module $\mathcal{L}_{\rm{TLA}}$ is the average of the two symmetrical losses.

\subsubsection{Concept-level Image-concept Alignment.} The ILA and TLA modules encode concepts based on the knowledge of the LLM, yet they come with certain limitations: (1) For the sake of efficiency, the parameters of the LLM are fixed in our framework; (2) We cannot guarantee that all knowledge derived from the LLM is entirely accurate. 
To alleviate these issues, we design a concept-level image-concept alignment module to fully exploit concept annotations directly sourced from the labels for more effective learning of the correlation and to further refine the alignment of attention-weighted image representations and concepts. To enhance the matching in the concept subspace, we initially make use of concept activation vectors (CAVs) to learn concept representations. 

Specifically, given concepts $C = \{c_1, c_2, ..., c_{N_c}\}$, where $c_i$ denotes the $i$-th concept (e.g., \textit{Blue-Whitish Veil}), $N_c$ denotes the number of concepts, we first split the concept samples $S^c = \{P^c, N^c\}$ into positive concept examples $P^c = \{P_i^c\}_{i=1}^{N_p}$ and negative concept examples $N^c = \{N_i^c\}_{i=1}^{N_n}$, where $P_i^c$ and $N_i^c$ are the CNN features of the images that contain or not contain the given concept $c$, respectively. $N_p$ and $N_n$ denote the number of positive and negative examples, respectively. We train an SVM to obtain the classification boundary, which separates features in $P^c$ and $N^c$. We learn the corresponding CAV $\bm b$ with weights $\omega^c$ and bias $\phi^c$, defined as the vector normal to the boundary, which satisfies $(\omega^c)^TP_i^c + \phi^c > 0$ and $(\omega^c)^TN_i^c + \phi^c < 0$ for all examples. Given the concept-based attention weighted image representation $g_i$ and the CAV $\bm b \in \mathbb{R}^{d \times N_c}$, where $d$ is the dimension of concept subspace, we project $g_i$ onto the concept subspace spanned by concept vectors using $b$:
\begin{equation}
    h_i = \frac{\langle g_i, \bm b\rangle}{||\bm b||^2}\bm b,
\end{equation}
where $h_i$ is the projected concept embedding, which can also be regarded as concept scores. Then, cross-entropy loss is applied to estimate the discrepancy between the concept scores and the concept ground truth:
\begin{equation}
    \mathcal{L_{\rm{CLA}}} = -\sum_{i=1}^N\left(C_i {\rm{log}}(h_i) + (1-C_i){\rm{log}}(1-h_i)\right)
\end{equation}

\definecolor{green2}{hsb}{0.5, 0.2, 1}
\renewcommand{\arraystretch}{1.5} 
\setlength{\tabcolsep}{5.8pt}  
\begin{table*}[t]  
\centering  
\fontsize{9}{9.5}\selectfont  
\begin{threeparttable}  
		  
\begin{tabular}{c|ccc|ccc|ccc}  
    \toprule\hline
    \multirow{2}{*}{\bf METHOD}&
    \multicolumn{3}{c|}{\bf Derm7pt}&
    \multicolumn{3}{c|}{\bf PH\textsuperscript{2}}&
    \multicolumn{3}{c}{\bf SkinCon}\cr
    &\bf AUC (\%) &\bf ACC (\%) &\bf F1 (\%) &\bf AUC (\%) &\bf ACC (\%) &\bf F1 (\%) &\bf AUC (\%) &\bf ACC (\%) &\bf F1 (\%) \cr
                
    \hline\hline 
    Sarkar et al. \shortcite{Sarkar_2022_CVPR}&76.22$_{2.06}$&73.89$_{1.47}$&66.81$_{1.23}$&79.33$_{0.62}$ &88.00$_{3.26}$ & 79.66$_{2.11}$ &68.21$_{1.44}$&71.14$_{1.21}$&71.32$_{1.38}$\cr
    PCBM \shortcite{yuksekgonul2023posthoc}&72.96$_{2.19}$&76.98$_{1.39}$&71.04$_{1.15}$&78.33$_{1.17}$&89.33$_{1.89}$ &81.49$_{2.57}$&68.94$_{1.59}$&71.04$_{1.13}$&70.47$_{0.75}$\cr  
    PCBM-h \shortcite{yuksekgonul2023posthoc}&83.27$_{1.14}$&79.89$_{0.89}$&74.48$_{1.37}$& 92.32$_{1.47}$&90.67$_{1.89}$&83.30$_{2.55}$& 69.53$_{1.67}$ & 72.28$_{1.39}$ & 72.28$_{1.29}$\cr  
    CBE \shortcite{Patricio_2023_CVPR}&76.60$_{0.35}$ & \underline{83.75$_{0.26}$}&\underline{78.13$_{0.44}$} &{97.50$_{0.00}$}&\underline{96.00$_{0.00}$}&93.89$_{0.00}$&72.75$_{1.15}$& 73.75$_{1.10}$ & 73.56$_{1.31}$\cr   
    \hline\hline 
    
    \rowcolor{green2!40} \bf MICA (w bot) & \underline{84.11$_{1.10}$} & {82.20$_{1.31}$}&  78.08$_{1.22}$ & \underline{97.66$_{1.24}$}&\underline{96.00$_{3.26}$}& \underline{94.40$_{1.48}$}& \underline{75.89$_{1.11}$}& \underline{74.29$_{1.09}$}&\underline{74.74$_{1.21}$}\cr
    \hline\hline
    \rowcolor{green2!40} \bf MICA (w/o bot)&{\bf 85.59$_{1.11}$}&{\bf 83.94$_{0.99}$}&\bf79.38$_{1.34}$&{\bf 98.18$_{1.43}$}&\bf 98.67$_{1.89}$&\bf 95.34$_{1.17}$&{\bf 75.92$_{1.13}$}&{\bf 75.63$_{1.07}$}&{\bf 75.43$_{1.24}$}\cr  
    \hline
   
\end{tabular}  
\end{threeparttable} 
\caption{Quantitative comparison on disease diagnosis with the concept-based state-of-the-arts. The performance is reported as mean$_{\rm std}$ of three random runs. Our method is highlighted in light cyan, where \textit{w bot} and \textit{w/o bot} represent with or without concept bottleneck, respectively. The best and second-best results are highlighted in \textbf{bold} and \underline{underlined}, respectively.}
\label{tab:diag_performance} 
\end{table*}

\noindent\textbf{Overall Objective} In the first stage of our method, we introduce Multi-Level Image-Concept Alignment modules to make full use of the concept annotations and jointly learn the correspondences between images and concepts. The overall training objective of the first stage is represented as:
\begin{equation}
    \mathcal{L} = \lambda_1 * \mathcal{L}_{\rm{ILA}} + \lambda_2 * \mathcal{L}_{\rm{TLA}} +  \lambda_3 * \mathcal{L}_{\rm{CLA}},
\end{equation}
where $\lambda_1$, $\lambda_2$, and $\lambda_3$ are hyperparameters.

\subsection{3.3 \quad Explainable Disease Diagnosis}

As illustrated in Figure \ref{overview}, the second stage of our method is to perform an explainable disease diagnosis. In order to mimic the process wherein medical experts first assess symptoms before making a final diagnosis during clinical treatments, we propose an explainable decision module with a concept bottleneck. Given the image encoder, which learned the image-concept association in the first stage, we initially utilize it to predict the clinical concepts in the images, i.e., detecting the presence, absence, and types of various symptoms within the images. Then, we apply a linear predictor that maps the concept subspace to the disease prediction based on the detected concepts. It is worth noting that the linear predictor is highly interpretable because its decision is based on the detected clinical concepts, which is consistent with human medical experts. In addition, the weight matrix of the linear predictor denotes the importance of each concept to the final decision. Moreover, a human medical expert can easily edit the predictor to get a more reliable diagnosis when observing a wrong or counter-intuitive phenomenon. Specifically, given the image representation $v_i$, of an input image $I_i$ encoded by the freezing image encoder $E_c$, we first detect the concepts and get the concept scores through an FC layer $f_c$, then the disease classification is performed based on the detected concepts through a linear layer $f_d$. 
We jointly train both the concept detection and disease classification layer through the following objective function:
\begin{equation}
    \begin{split}
    \hat{f}_c, \hat{f}_d = \mathop{\arg\min}_{{f}_c, {f}_d}\sum_{i=1}^N \left[CE(f_c(v_i), C_i) + \right.\\
    \left. \beta CE(f_d(f_c(v_i)), y_i)\right],
    \end{split}
\end{equation}
where $CE(\cdot)$ is the cross-entropy loss, $y_i$ is the diagnosis label of the $i$-th image, and $\beta$ is the hyperparameter to balance the concept detection and disease diagnosis.

\noindent\textbf{Discussion} What happens when we directly perform disease diagnosis using the trained image encoder without a concept bottleneck? Ideally, we would like to achieve higher performance while retaining the explainability of the concept-based methods. Thus, we simply apply a classification head on top of the image encoder to predict diagnosis labels. Since we only use the concept annotations when training the image encoder, it can be regarded as our model leveraging the concept knowledge to perform disease diagnosis, which is still an emulation of the $image \rightarrow concept \rightarrow diagnosis $ decision-making process.

\section{4 \quad Experiments}

\subsection{4.1 \quad Experimental Setup}

\subsubsection{Datasets:} \textbf{\textit{Derm7pt}} \cite{kawahara2018seven} is a dermoscopic image dataset contains 1011 images with clinical concepts
for melanoma skin lesions in dermatology. We consider all 7 dermoscopic concepts. Following \citet{Patricio_2023_CVPR}, we filter the dataset to obtain 827 images of \textit{Nevus} and \textit{Melanoma} classes. Only the dermoscopic images are used. The specific names of concepts can be found in section 4.2. 
\textbf{\textit{PH\textsuperscript{2}}} \cite{mendoncca2013ph} contains a total of 200 dermoscopic images of melanocytic lesions, including 80 common nevi, 80 atypical nevi, and 40 melanomas. 
We consider 5 concepts that are also included in the \textit{Derm7pt} dataset. Following previous work \cite{Patricio_2023_CVPR}, we combine the \textit{Common Nevi} and \textit{Atypical Nevi} classes of the \textit{PH\textsuperscript{2}} dataset into one global class label called \textit{Nevus}.
\textbf{\textit{SkinCon}} \cite{daneshjou2022skincon} is a skin disease dataset with 3230 images densely annotated by experts for fine-grained model debugging and analysis. 
We choose 22 concepts that have at least 50 images representing the concept from the \textit{F17k} \cite{17k} part. The classification categories are \textit{malignant}, \textit{benign} and \textit{non-neoplastic}. The dataset is split into training set, validation set and test set according to the proportion of 70\%, 15\% and 15\%, respectively.

\subsubsection{Compared Approaches:} 
\textbf{\citeauthor{Sarkar_2022_CVPR}} \shortcite{Sarkar_2022_CVPR} design an ante-hoc model where the output of the concept encoder is passed to a decoder that reconstructs the image, encouraging the model to capture the semantic features of the input image. \textbf{PCBM(-h)}  \cite{yuksekgonul2023posthoc} allows transfering concepts from other datasets and designs a residual modeling step to preserve performance of CBM. \textbf{CBE} \cite{Patricio_2023_CVPR} proposes a coherence loss to improve the visual coherence of concept activations. \textbf{CAV} \cite{lucieri2020interpretability} uses concept 
activation vectors to perform a detailed concept analysis for skin tumor classification.

\subsubsection{Implementation Details:} 
Our framework uses ResNet-50 \cite{resnet} as the image encoder. 
For concept encoder, we use a BERT encoder \cite{bert} with the ClinicalBERT weights \cite{alsentzer2019publicly}. During training in the first stage (i.e., multi-level alignment), we train the image encoder using only concept labels. In the second stage (i.e., disease diagnosis), the parameters of the image encoder are fixed, and we only train the classification heads. We adopt Adam \cite{kingma2014adam} optimizer with learning rate of 5e-5 in the first stage and 1e-4 in the second stage. For the hyperparameter selection, we use grid search and set $\tau_1 = 0.25$, $\tau_2 = 0.2$, $\tau_3 = 0.1$. We set $\beta = 1$ for \textit{Derm7pt} and \textit{SkinCon} dataset, and set $\beta = 0.5$ for \textit{PH\textsuperscript{2}} dataset.

\renewcommand{\arraystretch}{1.2} 
\begin{table}[t]  
\centering  
\fontsize{8}{10}\selectfont  
\begin{threeparttable}  
\begin{tabular}{ccccc}  
    \toprule\hline
    \bf Dataset&\bf Method&\bf AUC (\%)&\bf ACC (\%)&\bf F1 (\%) \\\hline
    
    \multirow{3}{*}{Derm7pt} 
    & CAV \shortcite{lucieri2020interpretability} &73.8&71.2&61.5\cr
    & CBE \shortcite{Patricio_2023_CVPR} &72.2&74.1&71.0\cr
    & \bf MICA (Ours) &\bf{78.6} & \bf{76.0} & \bf{72.6}\cr

    \hline
    
    \multirow{3}{*}{PH\textsuperscript{2}} 
    &CAV \shortcite{lucieri2020interpretability} &71.2&67.2&66.4\cr
    & CBE \shortcite{Patricio_2023_CVPR} &81.3&71.6&\bf{70.0}\cr
    & \bf MICA (Ours) & \bf{83.6} & \bf{75.2} & 68.4 \cr
    \hline

    \multirow{3}{*}{SkinCon} 
    & CAV \shortcite{lucieri2020interpretability} & 76.5 & 86.4 & 60.2\cr
    & CBE \shortcite{Patricio_2023_CVPR} &79.3&89.0&62.1\cr
    & \bf MICA (Ours) & \bf{82.6} & \bf{91.7} & \bf{63.8}\cr

    \hline

\end{tabular}  
\end{threeparttable}  
\caption{Quantitative results in concept detection.}
\label{tab:concept_detection} 
\end{table}

\renewcommand{\arraystretch}{1.4} 
\begin{table}[t]  
\setlength{\tabcolsep}{3pt}
\centering  
\fontsize{8}{10}\selectfont  
\begin{threeparttable}  
\begin{tabular}{ccc*{7}{c}}  
    \toprule\hline
    \bf Dataset&\bf Method&\bf PN&\bf DaG&\bf STR&\bf RS&\bf BWV&\bf PIG&\bf VS&\bf Avg.\\\hline
    
    \multirow{3}{*}{Derm7pt} 
    &CAV&72.7 &69.1 &74.3 &65.5 &74.6 &68.6&73.8&71.2\cr
    & CBE&\bf 78.5 &64.5 &\bf77.0 &71.0 &81.1 &\bf 75.4 &73.7&74.1\cr
    & \bf MICA& 74.4 & \bf 79.1 & 71.3 &\bf 72.8&\bf84.4& 68.8 & \bf81.3 & \bf 76.0\cr
    \hline
    \multirow{3}{*}{PH\textsuperscript{2}} 
    &CAV&68.0 &\bf60.0 &56.0 &76.0 & 76.0&N/A&N/A&67.2\cr
    &CBE&\bf 64.0 &58.0 & \bf 80.0&80.0 &76.0& N/A&N/A&71.6\cr
    &\bf MICA&60.0 &\bf60.0 & \bf 80.0 &\bf 88.0 & \bf 88.0 &N/A&N/A& \bf 75.2\cr
    
    \hline\bottomrule

\end{tabular}  
\end{threeparttable}  
\caption{Comparison of concept detection accuracy (\%). PN, DaG, etc. denote each clinical concept of the corresponding dataset and Avg. presents the average values of the previous columns. The best results are highlighted in bold.}
\label{tab:each_concept_detection} 
\end{table}  

\subsection{4.2 \quad Experimental Results}

To demonstrate that our method’s competitive performance on disease diagnosis and concept detection, we first compare with other state-of-the-art concept-based approaches on various datasets. Then we conduct comprehensive ablation experiments to validate the effectiveness of each module designed in our method. Finally, we evaluate the explainability of our method using multiple XAI metrics.

\subsubsection{Disease Diagnosis.} In Table \ref{tab:diag_performance}, we report the classification comparison results of our method under three metrics (AUROC, Accuracy and F1 Score) on the considered datasets. MICA outperforms other methods in overall performance especially AUC. As discussed in section 3.3, the results of MICA without concept bottleneck are also presented.

\begin{table}[th]  
\renewcommand{\arraystretch}{1.3} 
\setlength{\tabcolsep}{3pt}
\centering  
\fontsize{8}{10}\selectfont  
\begin{threeparttable}  
		  
\begin{tabular}{ccccccccc}  
    \toprule\hline
    \multirow{2}{*}{\bf ILA} & \multirow{2}{*}{\bf TLA} & \multirow{2}{*}{\bf CLA} & 
    \multicolumn{2}{c}{\bf Derm7pt}&
    \multicolumn{2}{c}{\bf PH\textsuperscript{2}}&
    \multicolumn{2}{c}{\bf SkinCon}\cr
    \cmidrule(lr){4-5}\cmidrule(lr){6-7}\cmidrule(lr){8-9}
    &&&AUC$_D$&AUC$_{C}$&AUC$_{D}$&AUC$_{C}$
    &AUC$_{D}$&AUC$_{C}$\\
    {\color{red}\XSolidBrush}&{\color{red}\XSolidBrush}&{\color{red}\XSolidBrush}
    & 71.2 & 63.1 & 88.3 & 70.6 & 67.4 & 70.7 \\
    {\color{green}\CheckmarkBold}&{\color{red}\XSolidBrush}&{\color{red}\XSolidBrush}
    & 81.1 & 75.5 & 95.0 & 80.5 & 72.9 & 78.5\\
    {\color{green}\CheckmarkBold}&{\color{green}\CheckmarkBold}&{\color{red}\XSolidBrush}
    & 82.8 & 77.3 & 95.6 & 84.5 & 74.7 & 81.5\\
    {\color{green}\CheckmarkBold}&{\color{red}\XSolidBrush}&{\color{green}\CheckmarkBold}
    & 83.3 & 78.0 & 96.4 & 82.7 & 75.1 & 81.8 \\
    {\color{red}\XSolidBrush}&{\color{green}\CheckmarkBold}&{\color{green}\CheckmarkBold}
    & 82.9 & 77.7 & 96.1 & 82.9 & 75.0 & 82.1 \\
    {\color{green}\CheckmarkBold}&{\color{green}\CheckmarkBold}&{\color{green}\CheckmarkBold}
    &\bf 84.1 &\bf 78.6 &\bf 97.7 &\bf 83.6 &\bf 75.9 &\bf 82.6 \\

    \hline\hline

\end{tabular}  
\end{threeparttable}  
\caption{Ablation study of MICA on disease diagnosis (AUC$_{D}$ [\%]) and concept detection (AUC$_{C}$ [\%]). ILA, TLA, and CLA represent the image-level, token-level, and concept-level alignment modules, respectively.}
\label{tab:ablation} 
\end{table}

\subsubsection{Concept Detection.} Table \ref{tab:concept_detection} shows the quantitative results of clinical concept detection. MICA outperforms other methods at most metrics in considered datasets. We also   report the test classification accuracy of each concept in \textit{Derm7pt} and \textit{PH\textsuperscript{2}} dataset in Table \ref{tab:each_concept_detection}. These two datasets have five common dermoscopic concepts from Seven-point Checklist \cite{7pt_checklist}, where PN stands for ``Pigment Network", DaG stands for ``Dots and Globules", STR stands for ``Streaks", RS stands for ``Regression Structures", BWV stands for ``Blue-Whitish Veil", PIG stands for ``Pigmentation", VS stands for ``Vascular Structures". Test accuracy of each concept reported is the average of the fine-grained labels that belong within each criteria. For example, PN (Pigment Network) denotes the mean accuracy of subclasses ``Typical Pigment Network" and ``Atypical Pigment Network". The ``avg" column reports the mean value of the test accuracy of all concepts.

\subsubsection{Ablation Study.} As shown in Table \ref{tab:ablation}, we observe that our framework can benefit from all three alignment modules, including ILA, TLA and CLA. The ablation results show that without any one of the three modules, the performance of both disease diagnosis and concept detection may suffer. The last configuration of Table \ref{tab:ablation} demonstrates that our method achieves the best overall performance with all three designed image-concept alignment components.

\begin{figure*}[t]
\centering
\subfigure[Intervention examples.]
{
    \label{intervene} 
    \includegraphics[width=0.52\columnwidth, height=3.5cm]{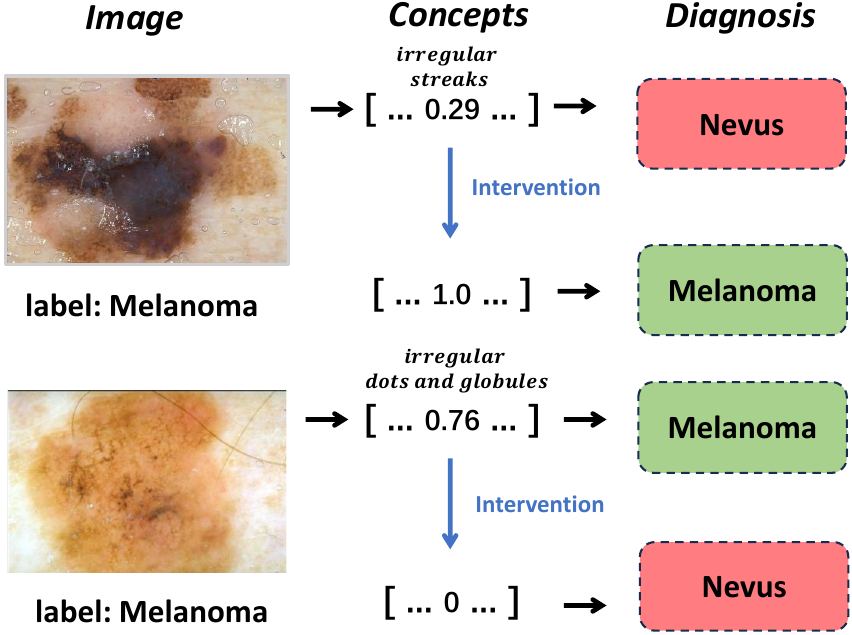} 
}
\ 
\subfigure[Intervention results.]
{   
    \label{intervene_result} 
    \includegraphics[width=0.51\columnwidth, height=3.5cm]{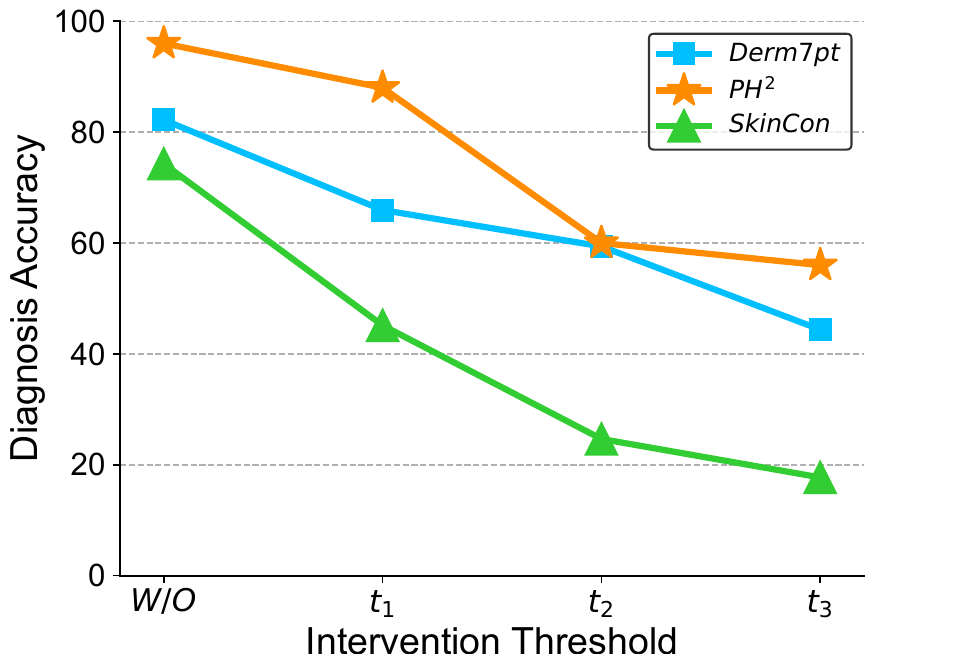} 
}
\ 
\subfigure[Visual and textual explanations.]
{
    \label{explanation}  
    \includegraphics[width=0.98\columnwidth, height=3.5cm]{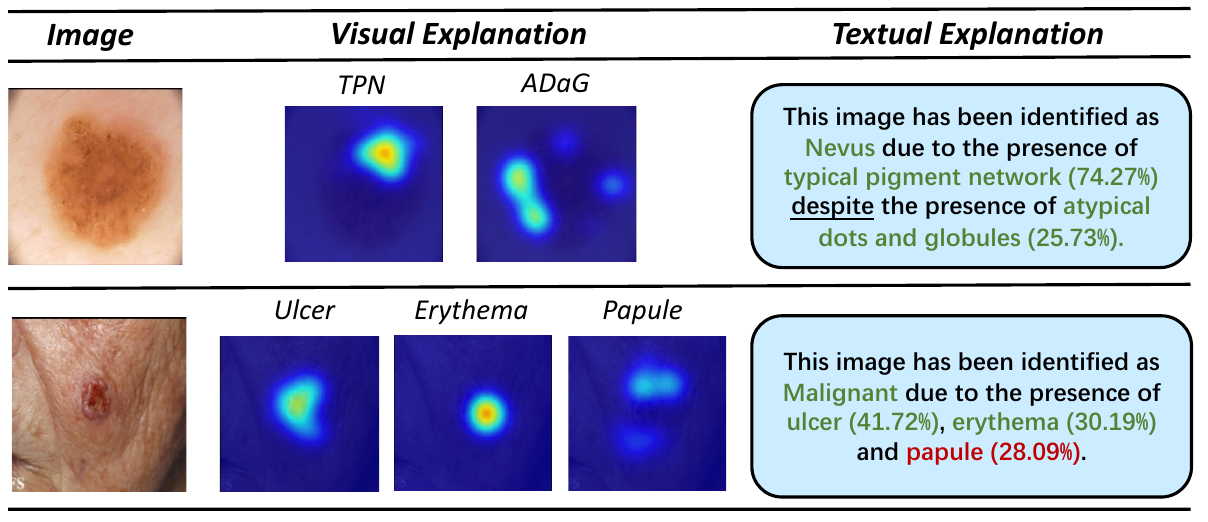} 
}
\caption{Illustration of our model's \textit{faithfulness}, \textit{plausibility} and \textit{understandability}. (a)(b) Test time concept-intervention examples and results. (c) Examples of visual and textual explanations provided by our method given skin images from different datasets. Correct prediction results are marked in green, while red highlights incorrect predictions.} 
\label{three_images}
\end{figure*}

\subsection{4.3 \quad Analysis of Explainability}
In this section, we evaluate and analyze the explainability of our method. Specifically, inspired by prior research \cite{hsiao2021roadmap,guidotti2018survey,rigotti2021attention,jin2023guidelines}, we outline and evaluate our framework on multiple essential metrics for XAI techniques, including \textit{faithfulness}, \textit{plausibility (understandability)} and \textit{efficiency}.

\subsubsection{Faithfulness.} \textit{Faithfulness} requires that the explanation should be highly faithful with the designed model mechanism and thus reflects the model decision process \cite{lakkaraju2019faithful}. In this paper, we employ test-time intervention on concepts to assess faithfulness. During inference time, we first predict the concepts and obtain the corresponding concept scores, then we intervene on a concept by changing its value. The diagnosis labels are predicted based on the concepts after intervention. Figure \ref{intervene} shows two cases of test-time intervention. In the first case, we replace the concept score with the ground-truth label of \textit{irregular streaks} and subsequently the diagnosis result changes from \textit{Nevus} to \textit{Melanoma}, which amends the diagnosis prediction. In the second case of Figure \ref{intervene}, we set the correct predicted value of concept \textit{irregular dots and globules} to 0 and the model decision changes from \textit{Melanoma} to \textit{Nevus}, which is consistent with the dermatologists' findings \cite{7pt_checklist,kawahara2018seven}. We also report test-time intervention results in Figure \ref{intervene_result}, which reflects the change of diagnosis accuracy when we zero those above the threshold values. The accuracy decreases when the threshold values decrease demonstrates that the predicted concepts are faithfully explaining the model's decisions.

\subsubsection{Plausibility \& Understandability.} \textit{Plausibility} refers to how believable or likely the given explanation seems, given what human-being know about the world or the domain of the problem \cite{carvalho2019machine,guidotti2018survey}, while \textit{understandability} refers to how easily a human user can comprehend the provided explanation without requiring technical knowledge \cite{jin2023guidelines}. In this paper, our model achieves both \textit{plausibility} and \textit{understandability} by providing concept-based visual and textual explanations. Figure \ref{explanation} shows the examples of explanations in detail. Given an input image, our method predicts the diagnosis labels with each predicted concept's localization and contribution (in \%). The concept localization is generated by visualizing the token-level correspondences with the images. As for the concept contribution, we leverage the softmax result of the last linear layer, which multiplies the concept logits and the corresponding weights. Our framework aggregates the prediction results to get textual explanations. It is worth noting that a ``despite" (underlined) is used in case of negative class fluence to signalize contradiction.

\subsubsection{Efficiency.} High label efficiency can allow the explainable model to be practically implemented in real-world applications without using extra data or annotations. In this paper, we assess our model's label efficiency by using different proportions of training data in the second stage. As shown in Table \ref{tab:label_efficient}, we report the disease diagnosis results using 100\%, 50\%, and 10\% of training data. For \textit{Derm7pt} and \textit{SkinCon} datasets, it can be observed that the diagnosis performance of the model does not exhibit significant decline when only 50\% or 10\% of the diagnosis labels are used. The obvious decrease for \textit{PH\textsuperscript{2}} dataset with 10\% training data may be because only 15 labels are used to train the classifier. Therefore, our method can achieve competitive diagnosis results only using a small proportion of diagnosis labels, signifying that our method encourages the model to learn the correspondences between medical images and clinical-related concepts, thus facilitating disease diagnosis effectively. This experimental observation is faithfully consistent with the doctors’ diagnostic process wherein medical experts make a diagnosis decision based on the detected symptoms.

\begin{table}[t]  
\renewcommand{\arraystretch}{1.3} 
\setlength{\tabcolsep}{5pt}
\centering  
\fontsize{8.5}{10}\selectfont  
\begin{threeparttable}  
		  
\begin{tabular}{ccccccc}  
    \toprule\hline
    \bf Dataset& 
    \multicolumn{2}{c}{\bf Derm7pt}&
    \multicolumn{2}{c}{\bf \ PH\textsuperscript{2}}&
    \multicolumn{2}{c}{\bf \ SkinCon}\cr
    \bf Num\# & 
    \multicolumn{2}{c}{34 / 173 / 346}& 
    \multicolumn{2}{c}{\ 15 / 75 / 150}&
    \multicolumn{2}{c}{\ 225 / 1126 / 2252}\cr
    \bf PCT (\%) &
    \multicolumn{2}{c}{10 / 50 / 100}& 
    \multicolumn{2}{c}{\ 10 / 50 / 100}&
    \multicolumn{2}{c}{\ 10 / 50 / 100}\cr
    \hline
    Proportion&AUC&ACC&\ AUC&\ ACC&\quad AUC&ACC\\
    10\% & 77.9 & 76.9 &\ 65.7 &\ 81.3 &\quad 70.1 & 71.5\\
    50\% & 83.2 & 81.0 &\ 96.5 &\ 93.3 &\quad 73.5 & 73.2\\
    100\% & 84.1 & 82.2 &\ 97.7 &\ 96.0 &\quad 75.9 & 74.3\\
                
    \hline\hline 
    
\end{tabular}  
\end{threeparttable}  
\caption{Disease diagnosis performance based on different portion of training data. PCT represents percentage.}
\label{tab:label_efficient} 
\end{table}

\section{5 \quad Conclusion}

In this paper, we propose MICA, a multi-modal explainable concept-based framework for skin disease diagnosis, which semantically aligns medical images and clinical-related concepts at multiple levels. By thoroughly learning the correspondences between images and concepts at the global image level, regional token level, and concept subspace level, our method outperforms other concept-based models while preserving inherent interpretability and offering both visual and textual explanations. Extensive experiments and explainability analysis conducted on skin image datasets demonstrate that our method simultaneously achieves superior performance, label efficiency, and interpretability.

\appendix

\section{Acknowledgments}
This work was supported by the Hong Kong Innovation and Technology Fund (Project No. ITS/028/21FP), Shenzhen Science and Technology Innovation Committee Fund (Project No. SGDX20210823103201011), and the Project of Hetao Shenzhen-Hong Kong Science and Technology Innovation Cooperation Zone (HZQB-KCZYB-2020083). 


\bibliography{aaai24}

\end{document}